\documentclass{article}

\usepackage[final,sglblindworkshop]{neurips_2025}
\workshoptitle{Machine Learning and the Physical Sciences}

\usepackage{graphicx} 
\usepackage{microtype}
\usepackage{graphicx}
\usepackage{amssymb}
\usepackage{mathtools}

\usepackage{amsthm}

\usepackage{amsmath}
\usepackage{subfigure}
\usepackage{booktabs}
 \usepackage{neurips_2025}
 \usepackage{wrapfig}
 \usepackage{subcaption}
\usepackage[utf8]{inputenc} 
\usepackage[T1]{fontenc}    
\usepackage{hyperref}       
\usepackage{url}            
\usepackage{booktabs}       
\usepackage{amsfonts}       
\usepackage{nicefrac}       
\usepackage{microtype}      
\usepackage{xcolor}   
\usepackage{neurips_2025}
\usepackage{graphicx} 
\usepackage{microtype}
\usepackage{graphicx}
\usepackage{subfigure}
\usepackage{booktabs} 

\usepackage[utf8]{inputenc} 
\usepackage[T1]{fontenc}    
\usepackage{hyperref}       
\usepackage{url}            
\usepackage{booktabs}       
\usepackage{amsfonts}       
\usepackage{nicefrac}       
\usepackage{microtype}      
\usepackage{xcolor}         

\title{Why Can't Neural Networks Master Extrapolation ? \\ Insights from Physical Laws }

\author{%
   Ramzi Dakhmouche  \\
{  \scriptsize{Institute of Mathematics, EPFL, Switzerland }}\\
 { \scriptsize Computational Engineering Lab, Empa, Switzerland} \\
\scriptsize  \texttt{ramzi.dakhmouche@epfl.ch} \\
  \And
 Hossein Gorji \\
  {\scriptsize{ Computational Engineering Lab, Empa, Switzerland }}
 \\
 \scriptsize \texttt{mohammadhossein.gorji@empa.ch}
}

\begin{document}

\maketitle

\begin{abstract}
Motivated by the remarkable success of Foundation Models (FMs) in language modeling, there has been growing interest in developing FMs for time series prediction, given the transformative power such models hold for science and engineering. This culminated in significant success of FMs in short-range forecasting settings. However, extrapolation or long-range forecasting remains elusive for FMs, which struggle to outperform even simple baselines. This contrasts with physical laws which have strong extrapolation properties, and raises the question of the fundamental difference between the structure of neural networks and physical laws. In this work,  we identify and formalize a fundamental property characterizing the ability of statistical learning models to predict more accurately outside of their training domain, hence explaining performance deterioration for deep learning models in extrapolation settings. In addition to a theoretical analysis, we present empirical results showcasing the implications of this property on current deep learning architectures. Our results not only clarify the root causes of the extrapolation gap but also suggest directions for designing next-generation forecasting models capable of mastering extrapolation. 
\end{abstract}

\section{Introduction}
In physics and engineering, the most impactful models are those that remain reliable even beyond the training data or observation domain, where controlled experiments or simulations are expensive, sparse, or beyond reach. Yet, the ability to master extrapolation remains out of reach for state-of-the-art machine learning models, including the latest forecasting approaches \cite{toner2025performance, liu2024moiraimoe, pucher2025evaluating, li2025tsfm} such as Foundation Models (FMs). FMs \cite{jin2024time, woo2024unified} are transformer-based neural networks with a considerably large number of parameters (hundreds of millions) pretrained  on large datasets from diverse time series domains, but still have been shown to be outperformed by simple linear or seasonal models \cite{toner2025performance} in extrapolation or long-range forecasting settings. More generally, deep learning models contrast with physical models having a symbolic structure, in the range of data regimes they can encode or generate. Indeed, in the context of fluid dynamics for instance, training a neural network solely on laminar flow data would lead to highly inaccurate predictions \cite{fukami2021model, sutter2021robust, chuang2020estimating} for turbulent flows. Whereas a symbolic differential equation such as the Navier-Stokes equation- or its discretization in space, represents a very flexible data-generating process that covers a very wide range of regimes. Furthermore, symbolic learning has demonstrated stronger extrapolation properties in biological applications \cite{dakhmouche2024robust, dakhmoucherobust}. This indicates a fundamental difference between symbolic models and neural networks and raises the question of characterizing this difference precisely to identify and design models that would extrapolate effectively. Although neural networks are typically over-parameterized and have a particular structure involving composing the same type of functions repeatedly, they are still approximators with explicit expressions, just like symbolic models, hence it is not clear what qualitative properties make them less suitable for extrapolation. Consequently, we tackle this question by making the following contributions:
\begin{itemize}
    \item We identify a precise characterization of structural variability as a key property that allows symbolic models to be better extrapolators than neural networks. 
    \item We propose a theoretical analysis demonstrating the role of this property in ensuring improved extrapolation under the Occam's razor hypothesis, thereby providing a principled basis for model selection outside the training range. 
    \item Building on this insight, we propose a minimal neural network architecture change as a first step toward better extrapolation and showcase its performance gains on synthetic and electricity time-series data.  
\end{itemize}

\section{Problem Setting}
\begin{wrapfigure}{r}{0.5\textwidth}  
    \includegraphics[scale=0.5]{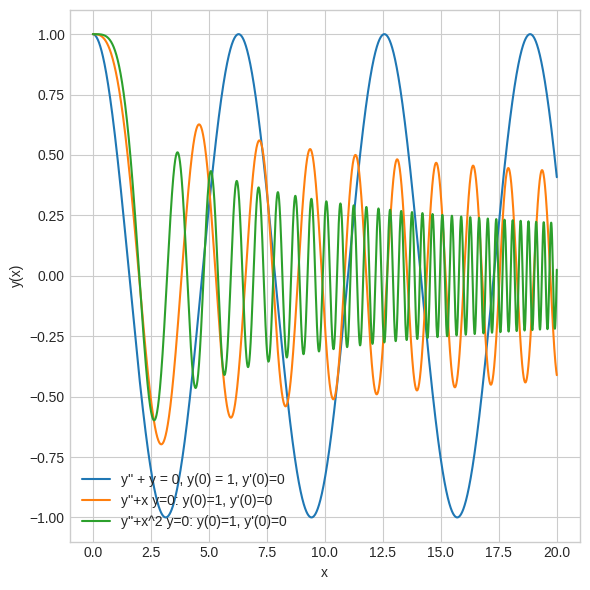}
    \caption{Information levels illustration. The green curve oscillates twice as much as the orange one, while they both have shrinking modes requiring more bits to be encoded. }
    \label{fig:sin}
\end{wrapfigure}
Given a dataset $(X_i, Y_i)_{i \leq n}$ of input and response variables $(x, y) \in \mathbb{R}^d \times \mathbb{R}^d$, we consider the regression task of predicting the response value $Y$ for new samples $X$. Assuming $(X_i)_{i \leq n}$ are sampled from a given domain $\mathcal{D} \subseteq \mathbb{R}^d$ with $Y_i = f(X_i)$ for all $i \leq n$, the goal in extrapolation is to achieve low prediction error on samples outside of $\mathcal{D}$. Note that, this is different from standard generalization in statistical learning theory, in that the latter focuses on ensuring low prediction error for new samples from the training domain or distribution. Importantly, extrapolation is an ill-posed problem in general, since outside of $\mathcal{D}$ the ground truth data generating function $f$ could have infinitely many different behaviors. Hence, to tackle this question in a meaningful way, we restrict the setting to cases where $f$ has an explicit expression. Such a set includes both neural networks and elementary functions such as polynomials, exponentials of polynomials, logarithms,  trigonometric functions \dots etc. One unifying property of all these functions is that they satisfy polynomial differential equations \cite{ritt1950differential, kolchin1973differential}. This allows us to define a selection principle based on Occam's razor \cite{udrescu2020ai}, by assigning preference to functions that encode less information. For a concise presentation, we assume $d = 1$. A relevant measure of information in this case is the number of bits needed to represent the polynomial ordinary differantial equation (ODE) satisfied by each function, and which is of the form $P(x, y', y'', y^{(3)}, \dots) = 0$ in the scalar input setting. This contrasts with previous pathwise or continuous information measures which are typically asymptotic \cite{cover1991entropy, bialek2001predictability}.  As illustrated in figure \ref{fig:sin}, functions corresponding to simpler ODEs encode less qualitative variation and hence correspond to simpler hypotheses. In the following, we therefore consider a model to yield better extrapolation, if it does so on simpler functions in this sense. 

\section{Measuring Structural Variability}
In order to ensure improved extrapolation, the key idea is to design a statistical learning model with structurally diverse building blocks, as we will analyze below. But first, how to measure structural variability ? Classical geometric notions such as dot product are too strict, while analytical measures such as norms are agnostic to structure. For instance, in $L^2(0, 2 \pi)$, trigonometric functions have dot product equal to 0 while having the same structure modulo a small shift (see figure \ref{fig:cos}). To overcome that, we propose to measure variability via algebraic objects derived from the ODEs satisfied by the model building blocks. Specifically, the measure has two components: 
\begin{wrapfigure}{r}{0.5\textwidth}  
    \includegraphics[scale=0.5]{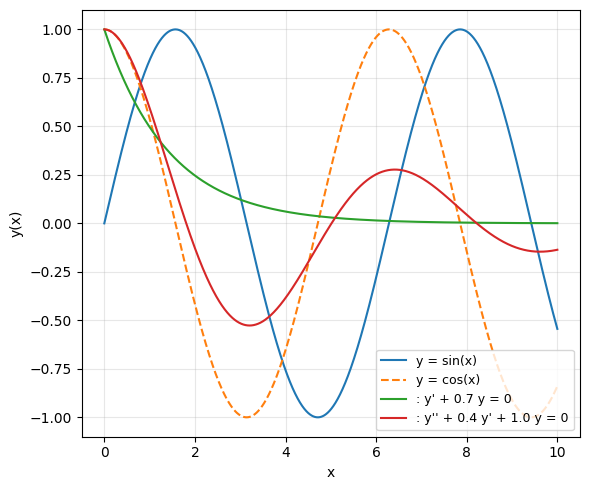}
    \caption{Structural variation illustration. The cosine function is just a shifted version of sine. }
    \label{fig:cos}
\end{wrapfigure}
\begin{enumerate}
    \item The order of the ODE: the highest order derivative of the ODE represents the most elementary discriminating quantity to consider, and encodes coarse qualitative change. For instance, oscillatory behavior cannot be observed in solutions of first order ODEs and require a second order term, as illustrated in figure \ref{fig:cos}. 
    \item Algebraic reduction classification: We focus here on polynomials of at most second degree, since the Navier-Stokes equation discretized in space is a quadratic ODE, yet captures regimes which are arguably as far apart as possible. More precisely, the linear case can be analyzed via the companion univariate annihilating polynomial defined for an ODE $y^{(n)} + c_n y^{(n-1)} + \dots + c_1 y = 0$ as $p(D) = D^n + c_n D^{n-1} + \dots + c_1 $,  where variability of the roots induces variability of the corresponding solutions. As for the quadratic case, it can be mapped to Sylvester’s Law of Inertia, which reduces second degree multivariate polynomials to sums of second degree monomials with coefficients in $\{-1, 0, 1 \}$. As illustrated in figure \ref{fig:law}, varying the choice of these coefficients for each term leads to a different  qualitative behavior for the ODE solutions. 
\end{enumerate}
Given the previous framework, we show in the following result that lacking a level of structural variability prevents accurate extrapolation. \\
~\\ \textbf{Proposition 1. } (Informal) \\ 
Consider a parameterized set of regression functions $\{ f_{\theta}: \mathbb{R} \longrightarrow  \mathbb{R} , \, \theta \in \mathbb{R}^p \} $ such that for $x \in  \mathbb{R}$
$$
f_{\theta} (x) = a_1 f_{1, \theta} (x) + \dots + a_k f_{k, \theta} (x)
$$
Assume $(f_{j, \theta})_{j \leq k}$ do not span the variability classes, then for all $M, \varepsilon>0$, there exists a smooth function $g_M$ such that 
$$
  \sum_{i=1}^n\|Y_i -g_M(X_i)\|_2 \leq \varepsilon \quad \textrm{and} \quad \inf_{\theta \in \mathbb{R}^p}  \| f_{\theta} - g_M \|_{\infty} > M 
$$
\emph{Proof.} We restrict the proof to the linear case and postpone it to Appendix B. \\
Next, we analyze the corresponding behavior for neural networks. 
\section{Differential Annihilators of Neural Networks} 
A natural question that arises given the proposed approach to measure structural variability is: what about the structure of the ODE satisfied by a neural network ? We provide an answer to this question in the following proposition, and will refer to such an equation as the differential annihilator.  \\ 
~\\
\textbf{Proposition 2.} (Informal) \\
Consider a neural network $f_{\theta}$ with Tanh or Sigmoid activation functions. Then, 
\begin{itemize}
    \item The minimal polynomial ODE satisfied by $f_{\theta}$ is of degree $\sum_{\ell}^L m_{\ell}$, where $L$ is its length and $m_{\ell}$ the width of each layer, and admits constant solutions which are highly dependent on the training data. 
    \item $f_{\theta}$ converges exponentially to a constant as it approaches the border of the training domain. 
\end{itemize}
Hence, the set of functions that a neural network can approach globally, that is, those whose differential annihilators can be well approximated by the annihilator of a neural network is considerably reduced, due to the constraint of constant solutions. We report the proof in Appendix C. 


\section{Numerical Results}
We evaluate our proposed framework by making a non-standard yet simple change in Multi-layer Perceptron (MLP) architecture to illustrate the gain obtained by structural variability. For that matter, we train a MLP that is a linear combination of subnetworks of varying length. This makes the order of the differential annihilators of each sub-network different. We use sigmoid activation functions and 3-layers with width of 16. We train with early stopping to prevent overfitting.  
\subsection{Synthetic Dataset}
\begin{figure}[h]
  \centering
  \begin{minipage}[b]{0.45\textwidth}
    \centering
    \includegraphics[width=\textwidth]{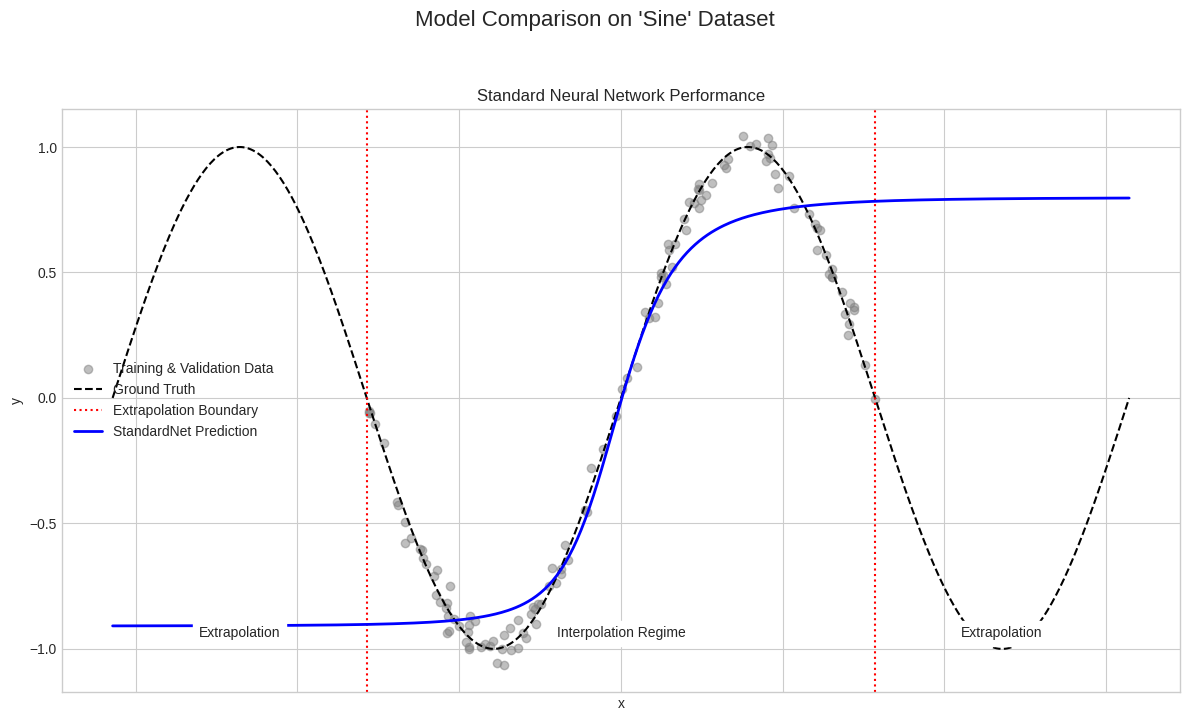}
    \caption{Predicted trajectory- Standard MLP}
    \label{fig:nv1bar}
  \end{minipage}
  \hfill
  \begin{minipage}[b]{0.45\textwidth}
    \centering
    \includegraphics[width=\textwidth]{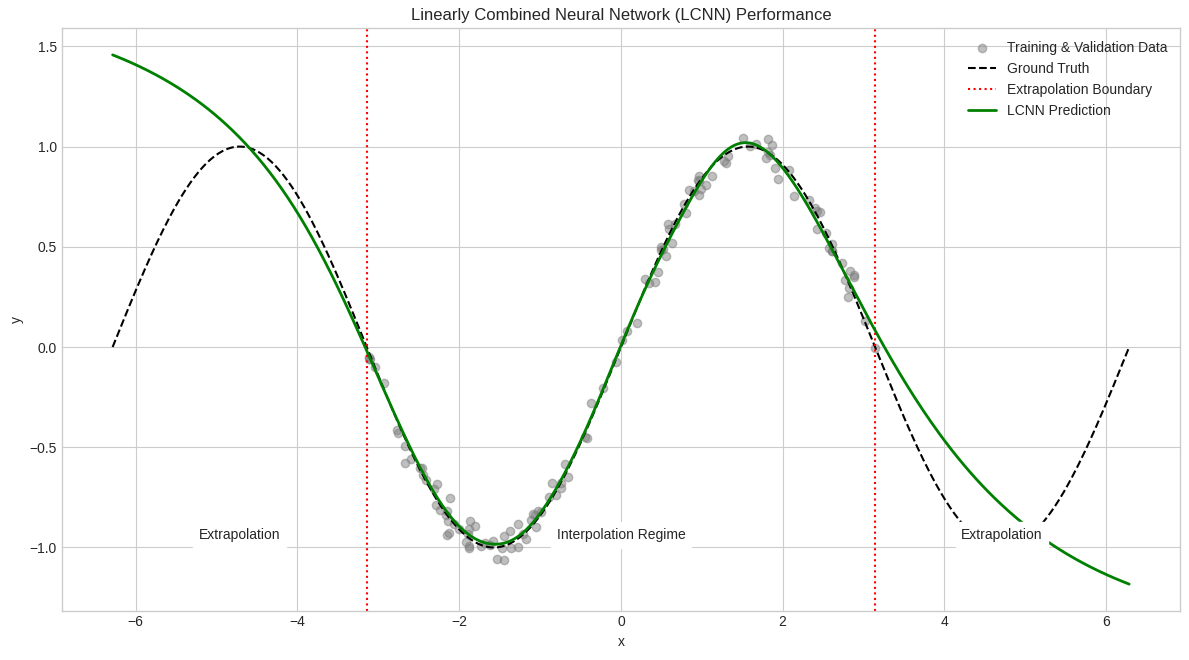}
    \caption{Predicted trajectory - Proposed MLP}
    \label{fig:nv2bar}
  \end{minipage}
  \end{figure}

We first evaluate the architectural change on elemantry function fitting, namely sine, complex periodic, quadratic and hyperbolic tangent. We report the extrapolation errors in table \ref{tab_elem} below. We report estimated trajectories for sin functions in figures \ref{fig:nv1bar} and \ref{fig:nv2bar}
\begin{table}[h]
\centering
\begin{tabular}{lcc}
\hline
Function & StandardNet (MSE) & Proposed Net (MSE) \\
\hline
Sin              & 1.172  & 0.082  \\
Complex Periodic & 1.592  & 0.416  \\
Quadratic        & 0.218  & 0.149  \\
Tanh             & 0.0018 & 0.0076 \\
\hline
\end{tabular}
\caption{Average extrapolation MSE for each function across window sizes.}
\label{tab_elem}
\end{table}

\subsection{Real-World Dataset}
We further evaluate on electric time series data from the ETTH dataset \cite{zhou2021informer, ilbert2024samformer}. We train on a window of 2500 samples and evaluate. We report the extrapolation results in table \ref{tab:extrap_mse_avg} below. Once again, we see an improvement in extrapolation accuracy via the proposed minimal architectural change. 

\begin{table}[ht]
\centering
\begin{tabular}{lcc}
\toprule
\textbf{Extrapolation Window size} & \textbf{StandardNet MSE} & \textbf{Proposed Net MSE} \\
\midrule
0.79 & 15.005008 & 11.537091 \\
1.57 & 14.166123 & 11.067607 \\
2.36 & 13.238490 & 11.514800 \\
3.14 & 12.184800 & 12.040763 \\
\bottomrule
\end{tabular}
\caption{Extrapolation MSE averaged over 3 runs.}
\label{tab:extrap_mse_avg}
\end{table}

\section{Conclusion}
In this work, we investigated the fundamental limitations of current neural networks for extrapolating beyond the training regime. By identifying and formalizing a key property that governs extrapolation performance, we provided both a theoretical explanation for the observed gap and empirical evidence demonstrating its impact.  Beyond clarifying the root causes of this phenomenon, our analysis points to concrete avenues for bridging the extrapolation gap, such as merging symbolic and over-parameterized models to increase structural variability. 

\section*{Acknowledgements }
This work was supported by the Swiss National Science Foundation under grant No. 212876. We acknowledge computational resources from the Swiss National Supercomputing Centre CSCS. R.D. acknowledges Dr. Ivan Lunati for providing laboratory infrastructure and computational resources and for valuable discussions.

\bibliography{Biblio}
\bibliographystyle{abbrv}

\newpage
\appendix

\section{Additional Numerical Results}
We report graphical representations of solutions of quadratic ODEs corresponding to various classes, featuring structural variability. 
\begin{figure}[h!]
    \centering
    \includegraphics[width=0.8\linewidth]{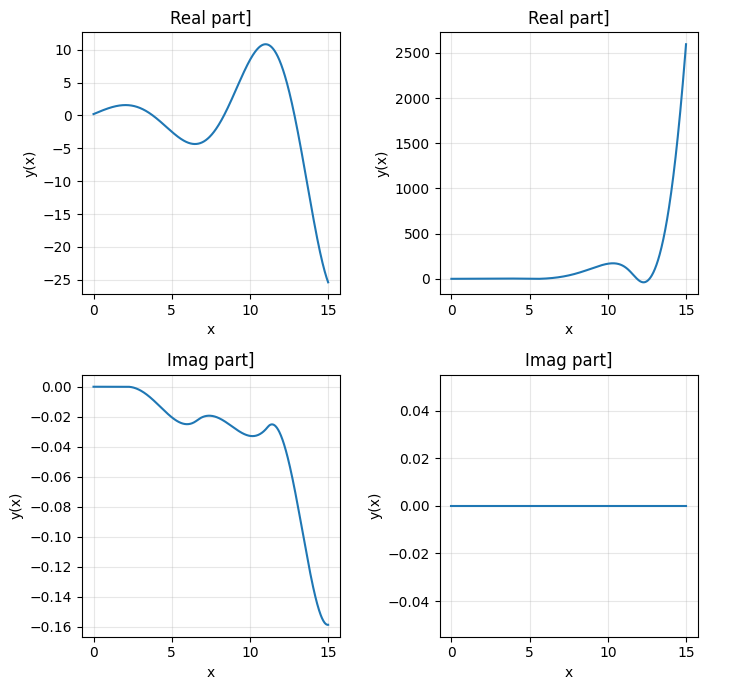}
        \caption{Structural variability illustration}
    \label{fig:law}
\end{figure}
\section{Proof of proposition 1}
\paragraph{Notation.}
Fix a compact interval \(K=[a,b]\subset\mathbb{R}\).
We denote by \(C(K)\) the Banach space of real–valued continuous functions on \(K\) with the sup–norm
\[
\|h\|_{\infty}:=\sup_{x\in K}|h(x)|.
\]
For \(r\in\mathbb{N}\), let \(C^r(K)\subset C(K)\) be the subspace of functions with \(r\) continuous derivatives on \(K\).

\paragraph{Model.}
Let \(\{f_{\theta}:\mathbb{R}\to\mathbb{R}:\theta\in\mathbb{R}^p\}\) be a parameterized family with a decomposition
\[
f_{\theta}(x)=\sum_{j=1}^k f_{j,\theta}(x),\qquad x\in\mathbb{R}.
\]

\paragraph{Variability deficit (precise).}
We say that the building blocks \(\{f_{j,\theta}\}\) \emph{do not span the variability classes} if there exists a nonzero constant–coefficient linear differential operator
\[
T=p(D)=D^r+c_rD^{r-1}+\cdots+c_1,\qquad r\ge 1,
\]
such that
\[
T f_{j,\theta}\equiv 0\ \text{on } K\quad\text{for all } \theta\in\mathbb{R}^p\ \text{and } j=1,\dots,k.
\]
Equivalently, every \(f_{\theta}\) lies in the closed linear subspace
\[
\mathcal{V}:=\ker(T)\cap C(K)=\{y\in C^r(K):Ty\equiv 0\ \text{on } K\}\subsetneq C(K).
\]

\paragraph{Proposition 1 (sup–norm separation under variability deficit).}
Assume the variability deficit above holds. Then for every $M, \varepsilon>0$, there exists a smooth function $g_M$ such that 
$$
  \sum_{i=1}^n\|Y_i -g_M(X_i)\|_2 \leq \varepsilon \quad \textrm{and} \quad \inf_{\theta \in \mathbb{R}^p}  \| f_{\theta} - g_M \|_{\infty} > M 
$$

\paragraph{Proof.}
Since \(T\not\equiv 0\), the subspace \(\mathcal{V}=\ker(T)\cap C(K)\) is a proper closed linear
subspace of \(C(K)\).
Choose \(g_0\in C^\infty(K)\setminus\mathcal{V}\) (e.g., any smooth function with \(Tg_0\not\equiv 0\) on \(K\)).

By the Hahn–Banach separation theorem in the Banach space \(C(K)\), there exists a bounded
linear functional \(\Lambda\in C(K)^*\) such that
\[
\Lambda|_{\mathcal{V}}=0
\quad\text{and}\quad
\Lambda(g_0)\neq 0.
\]
By the Riesz representation theorem for \(C(K)^*\), there is a finite signed Borel measure
\(\mu\) on \(K\) with
\[
\Lambda(h)=\int_K h(x)\,d\mu(x),\qquad \|\Lambda\|=\|\mu\|_{\mathrm{TV}}<\infty.
\]
For every \(\theta\) we have \(f_{\theta}\in\mathcal{V}\), hence \(\Lambda(f_{\theta})=0\). Therefore,
for all \(\theta\),
\[
|\Lambda(g_0-f_{\theta})|
=|\Lambda(g_0)|
\le \|\Lambda\|\,\|g_0-f_{\theta}\|_{\infty}.
\]
Taking the infimum over \(\theta\) yields
\[
\inf_{\theta}\|g_0-f_{\theta}\|_{\infty}
\;\ge\;\frac{|\Lambda(g_0)|}{\|\Lambda\|}\;=M+2>\;0.
\]
\smallskip
\noindent\textbf{Data correction.}
Let $\{x_i\}_{i=1}^n\subset K$ be the sample inputs. Choose pairwise disjoint open neighborhoods $U_i\Subset K$ and
$\phi_i\in C_c^\infty(U_i)$ with $\phi_i(x_i)=1$ and $\|\phi_i\|_\infty\le 1$.
Define coefficients
\[
t_i := \operatorname{clip}\big(y_i-g_0(x_i),\, \varepsilon/n\big),
\quad\text{i.e.}\quad
t_i=
\begin{cases}
\phantom{-}\varepsilon/n & \text{if } y_i-g_0(x_i) > \varepsilon/n,\\
-\varepsilon/n & \text{if } y_i-g_0(x_i) < -\varepsilon/n,\\
y_i-g_0(x_i) & \text{otherwise,}
\end{cases}
\]
and set the correction
\[
r(x):=\sum_{i=1}^n t_i\,\phi_i(x),\qquad g:=g_0+r.
\]
Because the supports are disjoint and $\|\phi_i\|_\infty\le 1$,
\[
\|r\|_\infty \;=\; \max_i |t_i| \;\le\; \frac{\varepsilon}{n} \;<\; 1 .
\]
At the data points we have $g(x_i)=g_0(x_i)+t_i$, hence
\[
|y_i-g(x_i)| = |y_i-g_0(x_i)-t_i| \le \frac{\varepsilon}{n},
\quad\Rightarrow\quad
\sum_{i=1}^n |y_i-g_0(x_i)| \le \varepsilon .
\]

\smallskip
\noindent\textbf{Margin to the model class.}
For any $s\in S:=\ker(T)$, by the triangle inequality,
\[
\|g-s\|_\infty \;\ge\; \|g_0-s\|_\infty - \|r\|_\infty.
\]
Taking the infimum over $s\in S$ and $\|r\|_\infty\le 1$,
\[
\mathrm{dist}(g,S)
\;\ge\;
\mathrm{dist}(g_0,S) - \|r\|_\infty
\;>\;
(M+2)-1
\;>\;
M,
\]
Since $\mathcal{F}\subset S$, it follows that
\[
\inf_{\theta}\|g-f_\theta\|_\infty \;\ge\; \mathrm{dist}(g,S) \;>\; M,
\]
which proves the claim.
\qed
\paragraph{Remarks.}
(1) The assumption “do not span the variability classes” is encoded here by the
existence of a nontrivial annihilating operator \(T\) that all building blocks satisfy on \(K\).
In the linear (constant–coefficient) case, \(\ker T\) is a finite–dimensional subspace
spanned by exponentials and sinusoidals, and the proof above is entirely within the sup–norm
on \(K\).

(2) The same argument applies to any situation in which the realized model class
\(\{f_{\theta}|_K:\theta\in\mathbb{R}^p\}\) is contained in a proper \emph{closed linear}
subspace of \(C(K)\).
In particular, even if one arrives at \(\mathcal{V}\) via quadratic jet constraints, it suffices
that the resulting feasible set be contained in some proper closed linear subspace \(V\subset C(K)\)
(e.g., by linearization on a restricted regime). The Hahn–Banach/Riesz separation then yields
the same conclusion.
\section{Proof of proposition 2}
\textbf{Proposition}[Polynomial ODE and boundary behavior for 1D$\to$1D MLPs with \texorpdfstring{$\tanh$}{tanh}/sigmoid] \\
Let $f_\theta:\mathbb R\to\mathbb R$ be a depth-$L$ multilayer perceptron with widths $(m_1,\dots,m_L)$, scalar input $x$, scalar output $f(x)$, and activation $\phi\in\{\tanh,\sigma\}$ (where $\sigma(u)=\frac1{1+e^{-u}}$). Set $M=\sum_{\ell=1}^L m_\ell$.

\begin{enumerate}
\item[(i)] There exists a nonzero polynomial $\mathcal P \in \mathbb R[T_0,\dots,T_M]$ (with coefficients depending on $\theta$) such that the scalar output satisfies the autonomous polynomial differential equation
\[
\mathcal P\big(f(x),f'(x),\dots,f^{(M)}(x)\big)\equiv 0.
\]
In particular, the minimal order of a polynomial ODE satisfied by $f_\theta$ is at most $M$. Moreover, this ODE admits constant solutions: there exist constants $c$ (depending on $\theta$) such that $f(x)\equiv c$ solves the ODE, i.e. $\mathcal P(c,0,\dots,0)=0$.

\item[(ii)] (Exponential convergence to constants at the boundary) Fix a bounded \emph{training domain} $[a,b]\subset\mathbb R$. Then there exist constants $f_\infty^-,f_\infty^+\in\mathbb R$ and positive constants $C_\pm,\kappa_\pm$ (depending on $\theta$) such that
\[
|f(x)-f_\infty^+|\le C_+ e^{-\kappa_+(x-b)}\quad (x\to +\infty),\qquad
|f(x)-f_\infty^-|\le C_- e^{-\kappa_-(a-x)}\quad (x\to -\infty).
\]
In words, as one moves beyond either border of the training interval, $f_\theta$ converges exponentially fast to a constant.
\end{enumerate}

\begin{proof}
\textbf{Network and notation.}
Write the usual forward equations
\[
h^{(0)}(x)=x,\qquad
z^{(\ell)}(x)=W^{(\ell)} h^{(\ell-1)}(x)+b^{(\ell)},\qquad
h^{(\ell)}(x)=\phi\!\big(z^{(\ell)}(x)\big)\in\mathbb R^{m_\ell},
\]
and $f(x)=\alpha^\top h^{(L)}(x)+\beta$. Stack all hidden activations into
\[
Y(x)\coloneqq\big(h^{(1)}(x),h^{(2)}(x),\dots,h^{(L)}(x)\big)\in\mathbb R^{M},
\]
and define the readout $G(Y)\coloneqq \alpha^\top h^{(L)}+\beta$, so $f(x)=G\!\big(Y(x)\big)$.

We shall use the elementary identities, valid for $y=\phi(u)$,
\[
\phi=\tanh:\quad \phi'(u)=1-\phi(u)^2=1-y^2,\qquad
\phi=\sigma:\quad \phi'(u)=\phi(u)\bigl(1-\phi(u)\bigr)=y(1-y).
\tag{$*$}
\]

\smallskip
\textbf{(i) Existence of a polynomial ODE of order $\le M$ and constant solutions.}

\emph{Step 1: An autonomous polynomial ODE for the hidden state.}
By the chain rule and ($*$), each first-layer neuron satisfies
\[
\frac{d}{dx}h^{(1)}_j(x)=w^{(1)}_j\,P\!\big(h^{(1)}_j(x)\big),
\]
with $P(y)=1-y^2$ (tanh) or $P(y)=y(1-y)$ (sigmoid). Thus the first-layer derivatives are \emph{polynomials in the first-layer activations}.

For layer $2$,
\[
\frac{d}{dx}h^{(2)}_k(x)
=\phi'\!\big(z^{(2)}_k(x)\big)\sum_{j} a_{kj}\,\frac{d}{dx}h^{(1)}_j(x)
=Q\!\big(h^{(2)}_k(x)\big)\sum_{j} a_{kj}\,w^{(1)}_j\,P\!\big(h^{(1)}_j(x)\big),
\]
where $Q(y)=1-y^2$ (tanh) or $Q(y)=y(1-y)$ (sigmoid). Hence second-layer derivatives are \emph{polynomials in $h^{(1)}$ and $h^{(2)}$}. Proceeding inductively over layers shows there exists a polynomial map
\[
F:\mathbb R^{M}\to\mathbb R^{M}\quad\text{such that}\quad
Y'(x)=F\!\big(Y(x)\big).
\tag{1}
\]

\emph{Step 2: $f^{(k)}$ are polynomial functions of $Y$.}
Define $H_0(Y)\coloneqq G(Y)$ and recursively
\[
H_{k+1}(Y)\coloneqq \nabla H_k(Y)\cdot F(Y)\qquad(k\ge 0).
\]
By construction and (1), $H_k$ is a polynomial for each $k$, and along the network trajectory $Y(x)$ we have
\[
f^{(k)}(x)=H_k\!\big(Y(x)\big),\qquad k=0,1,\dots.
\tag{2}
\]

\emph{Step 3: Algebraic elimination.}
Consider the polynomial map
\[
\Psi:\mathbb R^{M}\to\mathbb R^{M+1},\qquad
\Psi(Y)=\big(H_0(Y),H_1(Y),\dots,H_M(Y)\big).
\]
Its image is an algebraic set of dimension at most $M$. Hence there exists a nonzero polynomial
$\mathcal P\in\mathbb R[T_0,\dots,T_M]$ that vanishes on $\operatorname{Im}\Psi$ (e.g., by elimination/Nullstellensatz). Evaluating along $Y(x)$ and using (2) yields the autonomous polynomial differential equation
\[
\mathcal P\big(f(x),f'(x),\dots,f^{(M)}(x)\big)\equiv 0,
\]
establishing the order bound $\le M$.

\emph{Constant solutions.}
An equilibrium of (1) produces a constant solution of the $f$–ODE. To see that such equilibria exist, choose any vector $s^{(1)}\in S_\phi^{m_1}$ with
\[
S_\phi=\begin{cases}
\{\pm 1\}, & \phi=\tanh,\\
\{0,1\}, & \phi=\sigma,
\end{cases}
\quad\text{so that }P(s^{(1)}_j)=0\ \ \forall j.
\]
Then $h^{(1)}(x)\equiv s^{(1)}$ gives $h^{(1)}{}'(x)\equiv 0$, and by the layerwise formulae every higher-layer derivative also vanishes (they are linear combinations of lower-layer derivatives). Define recursively
\[
s^{(\ell)}\coloneqq \phi\!\big(W^{(\ell)} s^{(\ell-1)}+b^{(\ell)}\big),\qquad \ell=2,\dots,L,
\]
and set $\bar Y\coloneqq(s^{(1)},\dots,s^{(L)})$. Then $F(\bar Y)=0$, hence $H_0(\bar Y)=G(\bar Y)=:c$ and $H_k(\bar Y)=0$ for $k\ge 1$. Plugging into the identity $\mathcal P(H_0,\dots,H_M)\equiv 0$ gives
\[
\mathcal P(c,0,\dots,0)=0,
\]
so $f(x)\equiv c$ is a (constant) solution of the ODE. The set of such constants depends on $\theta$ through the affine maps and readout, hence is highly data/parameter dependent.

\smallskip
\textbf{(ii) Exponential convergence to constants outside a bounded domain.}

We prove the $x\to+\infty$ case; $x\to-\infty$ is analogous. For the first layer, each preactivation is $z^{(1)}_j(x)=w^{(1)}_j x+b^{(1)}_j$, so either $w^{(1)}_j=0$ (then $h^{(1)}_j$ is constant) or $|z^{(1)}_j(x)|\ge |w^{(1)}_j|x-|b^{(1)}_j|$ grows linearly with $x$.

Use the standard tails, for all $z\in\mathbb R$,
\[
\big|\tanh z-\operatorname{sgn}(z)\big|\le 2e^{-2|z|},\qquad
\big|\sigma z-\mathbf 1_{\{z>0\}}\big|\le e^{-|z|}.
\]
Therefore there exist constants $C_1,\kappa_1>0$ (depending on $w^{(1)},b^{(1)}$) and a saturation vector $s^{(1)}\in\{\pm1\}^{m_1}$ (tanh) or $s^{(1)}\in\{0,1\}^{m_1}$ (sigmoid) such that
\[
\|h^{(1)}(x)-s^{(1)}\|\le C_1 e^{-\kappa_1 x}\qquad (x\to+\infty).
\]

Inductively, write
\[
z^{(\ell)}(x)=W^{(\ell)} s^{(\ell-1)}+b^{(\ell)}\;+\;W^{(\ell)}\big(h^{(\ell-1)}(x)-s^{(\ell-1)}\big).
\]
The first two terms are constant; the last term is $O(e^{-\kappa_1 x})$. Since $\phi$ is globally Lipschitz with constant $\le 1$ for $\tanh$ and $\le 1/4$ for $\sigma$, there exist $C_\ell,\kappa_\ell>0$ and vectors $s^{(\ell)}=\phi\!\big(W^{(\ell)} s^{(\ell-1)}+b^{(\ell)}\big)$ such that
\[
\|h^{(\ell)}(x)-s^{(\ell)}\|\le C_\ell e^{-\kappa_\ell x}\qquad (x\to+\infty).
\]
Propagating to the linear readout gives
\[
|f(x)-f_\infty^+|\le C e^{-\kappa x}\qquad (x\to+\infty),
\]
with $f_\infty^+=\alpha^\top s^{(L)}+\beta$ and suitable $C,\kappa>0$ depending on $\theta$. The $x\to-\infty$ case yields $f_\infty^-$ similarly. Translating from $x\to\pm\infty$ to “distance beyond the border” $x-b$ or $a-x$ proves the stated estimates.
\end{proof}

\end{document}